\documentclass[10pt,twocolumn,letterpaper]{article}

\usepackage[accsupp]{axessibility}
\usepackage[pagenumbers]{cvpr} % To force page numbers, e.g. for an arXiv version

\usepackage{bbm}

\usepackage{algpseudocodex}
\usepackage{amsmath}
\usepackage{amssymb}
\usepackage{multirow}
\usepackage{booktabs}
\usepackage{makecell}
\usepackage{subcaption}
\usepackage{bbding}
\usepackage{graphicx}
\usepackage{caption}

\definecolor{cvprblue}{rgb}{0.21,0.49,0.74}
\usepackage[pagebackref,breaklinks,colorlinks,allcolors=cvprblue]{hyperref}

\definecolor{lightblue}{RGB}{68, 114, 196}

\def\paperID{14975} % *** Enter the Paper ID here
\def\confName{CVPR}
\def\confYear{2025}

\title{Exploring Scene Affinity for Semi-Supervised LiDAR Semantic Segmentation}

\author{\hspace*{-10pt}Chuandong Liu$^{1,\ast}$, \hspace*{4pt}
Xingxing Weng$^{1,\ast}$, \hspace*{4pt} 
Shuguo Jiang\textsuperscript{\rm 1 },  \hspace*{4pt}
Pengcheng Li\textsuperscript{\rm 3 },  \hspace*{4pt}
Lei Yu\textsuperscript{\rm 2,4 },  \hspace*{4pt}
Gui-Song Xia\textsuperscript{\rm 1,2,5,6 }\textsuperscript{\Envelope}  \\
    $^1$School of Computer Science, Wuhan University \\
    $^2$School of Artificial Intelligence, Wuhan University\\
    $^3$School of Computer Science and Technology, Hainan University \\
    $^4$School of Electronic Information, Wuhan University \\
    $^5$State Key Lab. of LIESMARS, Wuhan University\\
    $^6$Institute for Math \& AI, Wuhan University
}

\begin{document}
\maketitle
{\let\thefootnote\relax\footnotetext{{$^{\ast}$ Equal contribution.\ \textsuperscript{\Envelope} Corresponding authors.}}}

\begin{abstract}
This paper explores scene affinity (AIScene), namely intra-scene consistency and inter-scene correlation, for semi-supervised LiDAR semantic segmentation in driving scenes. Adopting teacher-student training, AIScene employs a teacher network to generate pseudo-labeled scenes from unlabeled data, which then supervise the student network's learning. Unlike most methods that include all points in pseudo-labeled scenes for forward propagation but only pseudo-labeled points for backpropagation, AIScene removes points without pseudo-labels, ensuring consistency in both forward and backward propagation within the scene. This simple point erasure strategy effectively prevents unsupervised, semantically ambiguous points (excluded in backpropagation) from affecting the learning of pseudo-labeled points. Moreover, AIScene incorporates patch-based data augmentation, mixing multiple scenes at both scene and instance levels. Compared to existing augmentation techniques that typically perform scene-level mixing between two scenes, our method enhances the semantic diversity of labeled (or pseudo-labeled) scenes, thereby improving the semi-supervised performance of segmentation models. Experiments show that AIScene outperforms previous methods on two popular benchmarks across four settings, achieving notable improvements of 1.9\% and 2.1\% in the most challenging 1\% labeled data. The code will be released at \href{https://github.com/azhuantou/AIScene}{https://github.com/azhuantou/AIScene}.
\end{abstract}

\begin{figure}[t]
    \centering
    \includegraphics[width=\linewidth]{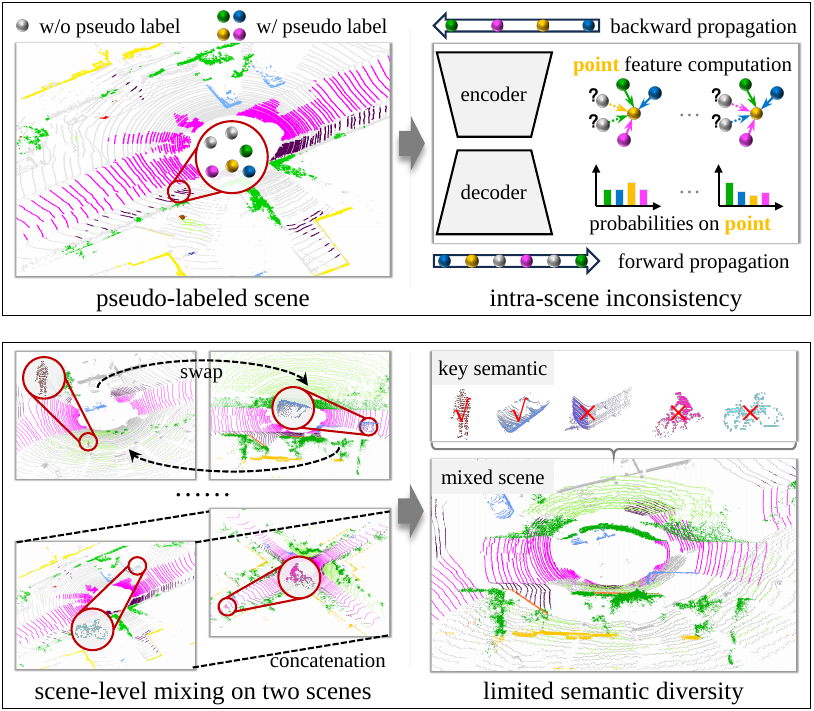}
    \caption{\textbf{Top:} Pseudo-labeled scenes generated from unlabeled scenes typically contain points with pseudo labels (colored) and without them (gray). Previous methods include all points in forward propagation but only pseudo-labeled points in backward propagation, leading to intra-scene inconsistency. This inconsistency may allow unsupervised, semantically ambiguous points to affect the learning of pseudo-labeled points (see \cref{fig:probability}). \textbf{Bottom:} Existing data augmentation techniques suitable for semi-supervised segmentation usually mix two scenes through scene-level operations like concatenation or swapping. Their resulting scenes may be constrained in terms of semantic diversity, potentially lacking objects like bicyclists and buses, which are important in driving scenes.}
    \vspace{-4mm}
    \label{fig:intro}
\end{figure}

\section{Introduction}
\label{sec:intro}
Semantic segmentation of LiDAR point clouds is essential for autonomous driving, allowing vehicles to perceive their 3D surroundings. Most state-of-the-art models for driving-scene LiDAR segmentation are dependent on high-quality, point-annotated datasets~\cite{hu2020randla,zhu2021cylindrical,su2023pups}. With numerous objects in driving scenes, densely annotating points is extremely labor-intensive. Consequently, semi-supervised LiDAR semantic segmentation has attracted growing attention, where models are trained on a limited amount of labeled data supplemented by a large set of unlabeled data.

The common solution for learning from unlabeled data is to apply teacher-student training, where a teacher network generates pseudo-labeled scenes from unlabeled data, which are then used to supervise the learning of a student network. The student network's objective is to minimize segmentation loss while the teacher network's weights are updated via exponential moving average~\cite{tarvainen2017mean}. To further improve model learning with unlabeled data, existing studies implement designs such as utilizing language-vision models~\cite{li2025lass3d}, adopting density-guided contrastive learning~\cite{li2024density}, or trying feature fusion between LiDAR and image data~\cite{chen2024beyond}. Despite the progress, two main issues remain to be addressed:

(1) Given the teacher network's confident prediction, pseudo-labeled scenes typically contain points with and without pseudo-labels. During training, most studies include all points in pseudo-labeled scenes in the student network's forward propagation, but only pseudo-labeled points are involved in backward propagation. Points excluded from backpropagation are unsupervised and, therefore, semantically ambiguous. However, these points still contribute to the forward feature computation for pseudo-labeled points, which may interfere with the model's ability to learn from them effectively. In short, the inconsistency between forward and backward propagation within the scene causes ambiguous points to obstruct the model's learning of pseudo-labeled points. Both point-based~\cite{hu2020randla} and voxel-based~\cite{choy20194d,zhu2021cylindrical} methods suffer from this issue in feature computation, as illustrated at the top of \cref{fig:intro}.

(2) Data augmentation is an effective practice in semi-supervised LiDAR semantic segmentation~\cite{lasermix,jiang2021guided}. The basic idea is to mix raw labeled scenes and pseudo-labeled scenes to enrich scene semantics. Although impressive advancements have been made~\cite{sheshappanavar2021patchaugment,zhang2022pointcutmix}, they usually select two scenes (one labeled scene and one pseudo-label scene) and then perform scene-level mix through operations like concatenation~\cite{nekrasov2021mix3d} and swapping~\cite{lasermix}, which limits the semantic diversity of mixed scenes. As shown at the bottom of \cref{fig:intro}, multiple scenes, whether labeled or pseudo-labeled, often share similar spatial layouts. Moreover, in driving scenes, objects such as cars and pedestrians are the primary focus of an autonomous vehicle's attention. This naturally raises the question: can we improve the semantic diversity of scenes by mixing multiple scenes at the instance level?

To address inconsistency and implement the multi-scene mixing idea, we explore \textbf{Scene} \textbf{A}ff\textbf{I}nity for semi-supervised LiDAR semantic segmentation, named AIScene. The term \textit{scene affinity} refers to intra-scene consistency and inter-scene correlation. Specifically, AIScene, built on the teacher-student training pipeline, employs a simple yet helpful point erasure strategy, removing points without pseudo-labels in forward propagation. This ensures intra-scene consistency in both forward and backward propagation, preventing semantically ambiguous points from disrupting learning from pseudo-labeled points. Building on this, AIScene incorporates patch-based data augmentation that mixes multiple scenes at both scene and instance levels. By maintaining two pools that store scene and instance patches from various scenes, it selects patches for mixing based on inter-scene correlation. This novel augmentation method not only enhances the semantic diversity of raw labeled scenes but also complements pseudo-labeled scenes where some points have been erased.

AIScene has been extensively evaluated through experiments and demonstrated superiority on two popular benchmarks for semi-supervised LiDAR semantic segmentation. It exceeds state-of-the-art methods by 1.3\% on the SemanticKITTI~\cite{kitti} and 1.2\% on the nuScenes~\cite{nuscene} in terms of average performance across four settings (1\%, 10\%, 20\%, and 50\% labeled data). Notably, in the most challenging setting, with only \textbf{1\% labeled data}, AIScene achieves remarkable improvements of \textbf{1.9\%} and \textbf{2.1\%} on the two datasets. Additionally, our ablation studies confirm AIScene's potential to reduce the requirements for labeled data across different LiDAR semantic segmentation networks (see \cref{tab:main_result} and ~\cref{tab:abl_main1}), with each component of AIScene able to function independently to facilitate semi-supervised LiDAR semantic segmentation (see~\cref{fig:consistency_multiscenes}).

To summarize, our contributions are as follows: 
\begin{itemize}
\item We propose a point erasure strategy to ensure intra-scene consistency and facilitate network learning from pseudo-labeled scenes. This strategy can serve as a plug-and-play tool for semi-supervised LiDAR semantic segmentation using the pseudo-label mechanism.

\item We design a patch-based data augmentation to mix multiple scenes at both scene and instance levels, guided by inter-scene correlation. This method improves the semantics diversity of labeled scenes and complements pseudo-labeled scenes.

\item Extensive experiments demonstrate that AIScene achieves state-of-the-art results on popular benchmarks for semi-supervised LiDAR semantic segmentation. Ablation studies show that the novel erasure strategy and augmentation method can function independently.
\end{itemize}

\begin{figure*}[!thp]
    \centering
    \includegraphics[width=17cm]{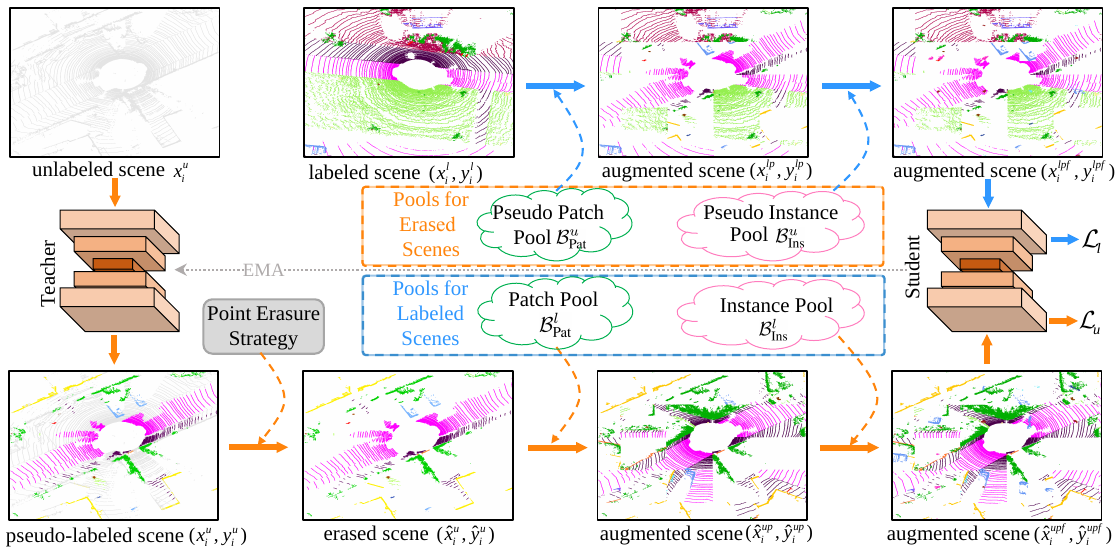}
    \caption{Overview of the proposed AIScene training pipeline. The teacher network generates the pseudo-labels $y_i^u$ from the unlabeled scene $x_i^u$, followed by an erasure operation to filter out low-confidence points, ensuring training consistency. For the labeled scene $x_i^l$ and the generated erased scene $\hat{x}_i^{u}$, we employ patch-based data augmentation to obtain augmented scenes $x_i^{lpf}$ and $\hat{x}_i^{upf}$, respectively. Subsequently, both scenes are fed into the student network to compute the loss of $\mathcal{L}_l$ and $\mathcal{L}_u$.}
    \vspace{-4mm}
    \label{fig:pipline} 
\end{figure*} 

\section{Related Work}
\textbf{Fully-supervised LiDAR Semantic Segmentation:}
Numerous approaches for LiDAR semantic segmentation have been explored, which can be broadly grouped into three categories: voxel-based, 2D projection-based, and point-based methods. Voxel-based approaches~\cite{cheng20212,choy20194d,tang2020searching,zhu2021cylindrical,lai2023spherical} involve voxelizing sparse point clouds to generate regular voxel grids, allowing the application of standard 3D convolution neural networks for embedding learning. Conversely, 2D projection-based methods~\cite{kong2023rethinking,milioto2019rangenet++,wu2019squeezesegv2,xu2020squeezesegv3,zhang2020polarnet,zhao2021fidnet} project input point clouds into a 2D dense space, use 2D convolutional filters~\cite{ronneberger2015u} for processing, and then map predictions back to the original 3D space. Finally, point-based methods~\cite{hu2020randla,qi2017pointnet,lai2022stratified,wu2024point} operate directly on raw point clouds, enabling the extraction of point-wise semantic features.

\noindent \textbf{Semi-supervised LiDAR Semantic Segmentation:}
To reduce the need for 3D dense annotations, recent research has explored methods that require less supervision~\cite{kweon2024weakly,liu2022ss3d,scribble,zhang2021weakly,chen2023clip2scene,tang2024all}.
Semi-supervised learning~\cite{lessismore,reichardt2023360deg,zou2018unsupervised,chen2021semi,li2024density,lasermix++} aims to utilize a large number of unlabeled data to alleviate the annotation burden.
LaserMix~\cite{lasermix}, a pioneer work in semi-supervised LiDAR segmentation, leverages a simple laser beam mix strategy to achieve solid results.
Subsequent research fuses LiDAR and camera data to extract deeper semantic information~\cite{chen2024beyond,ignet}. DDSemi~\cite{li2024density}, focusing on point cloud density, builds on contrastive learning to enhance feature discriminability. The recently proposed LASS3D \cite{li2025lass3d} leverages large language-vision models to support semi-supervised segmentation. Existing semi-supervised methods often build upon the mean-teacher paradigm~\cite{tarvainen2017mean} and utilize pseudo labels~\cite{pseudo_label_1}. However, many points without pseudo-labels contribute equally to pseudo-labeled points in forward propagation but are ignored during backward propagation, leading to intra-scene inconsistency. To address this, our AIScene leverages a point erasure strategy to mitigate this inconsistency, complementing the pseudo-label mechanism.

\noindent \textbf{3D Point Cloud Data Augmentation:}
Point cloud data augmentation has been explored across various 3D vision tasks~\cite{aug_survey,shin2023diversified,wang2024pointpatchmix,fang2021lidar,wang2023ssda3d,liu2023hierarchical}.
Inspired by the success of Mixup~\cite{zhang2017mixup} in image classification, several methods have investigated techniques for mixing selected point cloud samples. To preserve structural information, RSMix~\cite{lee2021regularization} introduces a neighboring function, and PointCutMix~\cite{zhang2022pointcutmix} establishes correspondences for sample fusion. Notably, these methods focus on local object-level augmentation.
In response, several works have explored mixing strategies for point cloud scene augmentation. PloarMix~\cite{xiao2022polarmix} swaps points within the fan-shaped areas between two scenes, while CoSMix~\cite{saltori2023compositional} aims to mitigate domain shift by mixing semantic point groups.
Unlike these methods, our method is capable of modeling inter-scene information across multiple scenes to create more diversified data, while also extracting instance-level semantic information for individual classes to further enrich scenes.

\section{Method}
\subsection{Preliminary}
\label{sec:3.2}
\textbf{Problem Formulation.}
Before we describe our model and training process, we first provide the definition of semi-supervised point cloud segmentation.
In detail, we have a small labeled training set $D_{l}=\left\{x_{i}^{l}, y_{i}^{l}\right\}_{i=1}^{N_{l}}$ and a large unlabeled training set $D_{u}=\left\{x_{i}^{u}\right\}_{i=1}^{N_{u}}$, where $x_{i} \in \mathbb{R}^{m \times {(3+r)}}$ is a point cloud scene of $m$ points with three-dimensional Cartesian coordinates relative to the sensor and additional $r$-dimensional feature (\eg, intensity, color, or reflectivity), $y_{i}$ represents its corresponding point-level segmentation labels, $N_{l}$ and $N_{u}$ are the numbers of labeled and unlabeled point cloud scenes, respectively, and $N_{u} \gg N_{l}$ in ideal cases.
The goal of SSL-based point cloud segmentation is to learn a segmentation model using limited labeled data and large-scale unlabeled data.

\noindent\textbf{Teacher-Student Paradigm.}
Following current mainstream research \cite{lasermix,li2024density}, our framework also builds upon the teacher-student paradigm, which includes two segmentation networks with the same architecture.
Our AIScene can be instantiated with any 3D segmentation network~\cite{zhu2021cylindrical,choy20194d,lai2023spherical}, denoted as $\Phi(\cdot)$.
In detail, the teacher model generates pseudo-labels for unlabeled scenes and undergoes a gradual update through the exponential moving average (EMA)~\cite{tarvainen2017mean} of the student weights:
\begin{align} \label{ema}
        \theta_\mathbbm{t}^{k+1}=\alpha \cdot \theta_\mathbbm{t}^{k} + (1-\alpha) \cdot \theta_\mathbbm{s}^{k},
\end{align}
where $\alpha$ is the EMA decay rate, which is set to 0.99 by default, $\theta_\mathbbm{t}$ and $\theta_\mathbbm{s}$ represent the weight parameters of the teacher and student models, respectively, and $k$ denotes the training step.

\subsection{Point Erasure Strategy}
Based on the teacher-student mutual learning framework~\cite{sohn2020fixmatch}, the overall architecture of AIScene is depicted in \cref{fig:pipline}. Specifically, AIScene consists of two primary branches: one dedicated to labeled scenes and the other to unlabeled scenes. How to effectively leverage a large number of unlabeled scenes is crucial for semi-supervised learning. Hence, for the branch of unlabeled scenes, we use a teacher network to generate pseudo-labeled scenes $(x_i^u, y_i^u)$ based on a predefined confidence threshold $\tau_s$~\cite{lasermix}. As mentioned previously, when using the pseudo-labeled scene for training, points without pseudo-labels are involved in the forward propagation but are ignored in the backward propagation due to the lack of corresponding supervision. This leads to intra-scene inconsistency, which causes semantic ambiguity of points. To address this issue, we propose erasing points without pseudo labels, resulting in an erased scene $(\hat{x}_i^u, \hat{y}_i^u)$ that contains fully annotated information, similar to fully supervised scenes. Formally, intra-scene inconsistency addressed by the point erasure strategy can be defined as:
\begin{align} \label{erasing_scene}
	\hat{x}_i^u = \{ x_i^u | \Phi_s(x_i^u) \geq \tau_s\}, 
\end{align}
where $\hat{x}_i^u$ is the erased scene with corresponding class labels $\hat{y}_i^u$, $\Phi_s(\cdot)$ represents the segmentation network that predicts the confidence score for each point in the scene $x_i^u$, and $\tau_s$ is a predefined confidence threshold, equivalent to the pseudo-label mining threshold, requiring no additional adjustment. Since the proposed point erasure strategy is independent of the segmentation network architecture, it can be utilized as a plug-and-play tool for semi-supervised 3D semantic segmentation methods based on the pseudo-label mechanism, as depicted in \cref{fig:consistency_multiscenes} (a).

\begin{figure}[t]
    \centering
    \includegraphics[width=8.2cm]{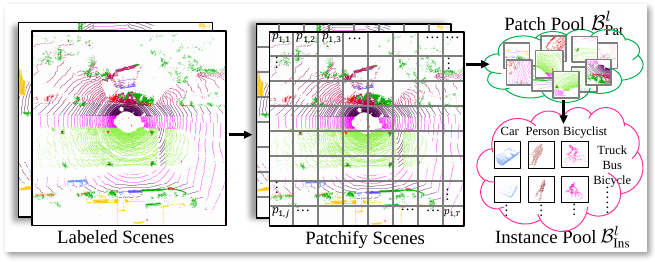}
    \caption{Illustration of the pool initialization process. For each labeled point cloud scene, splitting it into multiple patches from the BEV perspective and assigning different instance categories to each patch enables the extraction of instances belonging to different classes. Subsequently, these patches and instances are stored individually in the patch pool and instance pool. 
    }
    \label{fig:pool_init}
\end{figure}

\subsection{Patch-based Data Augmentation}
Based on the labeled and erased scenes, we propose a patch-based data augmentation to enhance the semantic diversity of annotation information, guided by inter-scene correlation. The key idea is to construct patch pools to store scene-level and instance-level information, and then combine patches from multiple scenes to generate diverse training samples.

\textbf{Pool Initialization.} We first use the labeled training set $D_{l}$ to create two patch pools, which store scene-level and instance-level information, respectively, as shown in \cref{fig:pool_init}. Given the $i$-th point cloud scene $x_{i}^{l}$, we patchify $x_{i}^{l}$ into a sequence of $T$ regular, non-overlapping point patches $\{p_{ij}^l\}_{j=1}^{T}$, each of size $(n \times n)$, in the bird's eye view. This process can be expressed as:
\begin{align} \label{labeled_scene}
    \left\{p_{ij}^l\right\}_{j=1}^{T} = \text{patchify} (x_{i}^{l}),
\end{align}
where $p_{ij}^l$ denotes the $j$-th point patch from the sequence. For all labeled scenes, by perform the operation in \cref{labeled_scene}, the patch pool $\mathcal{B}_\text{Pat}^l$ is constructed, which can be defined as:
\begin{align} \label{patch_pool1}
    \mathcal{B}_\text{Pat}^l = \left\{ \left\{p_{ij}^l\right\}_{j=1}^{T} \right\}_{i=1}^{N_l}.
\end{align}
Based on the patch pool, we can obtain the instance pool $\mathcal{B}_{\text{Ins}}^l$ by assigning different class labels to each patch, as depicted in \cref{fig:pool_init}, which contains instance information for different categories (\eg, car and person). 

Following the same steps used for labeled scenes, we further create a pseudo patch pool $\mathcal{B}_{\text{Pat}}^u$ and a pseudo instance pool $\mathcal{B}_{\text{Ins}}^u$ from the erased scene $\hat{x}_i^u$. Unlike $\mathcal{B}_\text{Pat}^l$ and $\mathcal{B}_\text{Ins}^l$ covering the whole training set $D_{l}$, the pseudo pools $\mathcal{B}_\text{Pat}^u$ and $\mathcal{B}_\text{Ins}^u$ only contain scene and instance information from the current batch and are reset in each iteration. This prevents low-quality pseudo labels from persisting in pools and misleading model training. Additionally, if the number of points within a patch or instance falls below a certain threshold $\tau_{\text{min}}$, it will not be saved into the pool. 
In practice, we set $\tau_{\text{min}}$ to 5~\cite{yan2018second}. If a patch or instance has fewer than 5 points, its spatial structure exhibits high uncertainty, making it semantically ambiguous. Thus, it is excluded from the pools to prevent noise during data augmentation.

\begin{figure}[t]
    \centering
    \includegraphics[width=8.5cm]{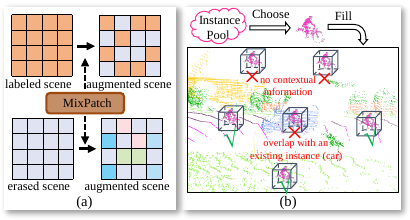}
    \caption{
    (a) Illustration of the MixPatch process. (b) Illustration of selecting instances from the patch pool for filling. An instance will not be filled if there are no other points around the filling location or if the filled instance overlaps with existing instances.
    }
    \label{fig:mixpatch}
\end{figure}

\textbf{MixPatch.} To model the inter-scene correlation and then guide the transfer of annotation information between labeled scenes and unlabeled scenes, we combine multiple scenes using patch pools. \cref{fig:mixpatch} (a) illustrates the patch mixing (MixPatch) strategy, while \cref{fig:mixpatch} (b) presents the instance filling (InsFill) strategy, which is introduced in the next paragraph. During a training iteration, the mini-batch is comprised of labeled and unlabeled scenes. After the unlabeled scene $x_i^u$ is transformed into the erased scene $\hat{x}_i^u$ using the point erasure strategy, we integrate information from the patch pool into $\hat{x}_i^u$ by MixPatch. The MixPatch process can be described as follows:
\begin{align} \label{clean_scene_patch}
	\hat{x}_i^{up} = (\hat{x}_i^{u} \odot \mathcal{M}) \oplus ( \mathcal{B}_\text{Pat}^l \odot (1 - \mathcal{M}) ),
\end{align}
where $\mathcal{M}$ is a binary mask indicating which patches from $\hat{x}_i^{u}$ will be replaced with selected patches of the same patch index from the patch pool, $\odot$ denotes element-wise multiplication, and $\hat{x}_i^{up}$ represents the augmented scene.
For labeled scenes, MixPatch is applied based on the current pseudo patch pool $\mathcal{B}_\text{Pat}^u$, which can be expressed as the following:
\begin{align} \label{labeled_scene_patch}
	x_i^{lp} = (x_i^{l} \odot (1 - \mathcal{M})) \oplus ( \mathcal{B}_\text{Pat}^u \odot \mathcal{M} ),
\end{align}
where $\mathcal{M}$ is the same as defined in \cref{clean_scene_patch}, thereby ensuring the full utilization of scene information from the unlabeled scenes. The binary mask $\mathcal{M}$ and its complement $1 - \mathcal{M}$ are used to merge inter-scene point cloud patches with the erased scene, facilitating the generation of diverse semantic information for labeled and unlabeled scenes. To alleviate bias in the patch mixing process, we apply a uniform sampling strategy to the patches in each scene, ensuring that each patch has an equal opportunity of being selected in the mixed scene.

\textbf{InsFill.} In addition to patch mixing, we utilize the instances stored in the instance pool $\mathcal{B}_\text{Ins}^l$ to further enhance the scene $\hat{x}_i^{up}$. Given a patch with location index $j$ in the scene, we randomly select an instance from the instance pool $\mathcal{B}_\text{Ins}^l$ that was extracted from the $j$-th patch. Patches with identical indices typically correspond to similar LiDAR spatial layouts, ensuring consistency between the selected instance and its fill positions in the scene. As depicted in \cref{fig:mixpatch} (b), if the selected instance overlaps with existing instances in the scene or lacks sufficient contextual information, the filling operation is bypassed. This judgment process can be easily implemented using the maximum 3D bounding boxes. This data augmentation provides instance-level semantic information for the point cloud scene, enabling the augmented scene $\hat{x}_i^{upf}$ and corresponding labels $\hat{y}_i^{upf}$ to effectively support semi-supervised training. For the labeled branch, we also apply the same filling strategy as described, selecting instances from the pseudo instance pool $\mathcal{B}_\text{Ins}^u$ to fill the current scene. This InsFill process yields the scene $x_i^{lpf}$ along with its corresponding label $y_i^{lpf}$.

\begin{table*}[t]
\centering
\resizebox{17cm}{!}{
\begin{tabular}{c|c|cccc|c|cccc|c}
\hline
\multirow{2}{*}{Method} & \multirow{2}{*}{Venue} & \multicolumn{4}{c|}{SemanticKITTI \cite{kitti}} & \multirow{2}{*}{Avg.} & \multicolumn{4}{c|}{nuScenes \cite{nuscene}} & \multirow{2}{*}{Avg.} \\ \cline{3-6} \cline{8-11}
 &  & 1\% & 10\% & 20\% & 50\% &  & 1\% & 10\% & 20\% & 50\% &  \\ \hline \hline
Meanteacher \cite{tarvainen2017mean} & NeurIPS'2017 & 45.4 & 57.1 & 59.2 & 60.0 & 55.4 & 51.6 & 66.0 & 67.1 & 71.7 & 64.1 \\
CBST \cite{zou2018unsupervised} & ECCV'2018 & 48.8 & 58.3 & 59.4 & 59.7 & 56.6 & 53.0 & 66.5 & 69.6 & 71.6 & 65.2 \\
CPS \cite{chen2021semi} & CVPR'2021 & 46.7 & 58.7 & 59.6 & 60.5 & 56.4 & 52.9 & 66.3 & 70.0 & 72.5 & 65.4 \\ \hline
LaserMix (Range View) \cite{lasermix} & CVPR'2023 & 43.4 & 58.8 & 59.4 & 61.4 & 55.8 & 49.5 & 68.2 & 70.6 & 73.0 & 65.3 \\
LaserMix (Voxel) \cite{lasermix} & CVPR'2023 & 50.6 & 60.0 & 61.9 & 62.3 & 58.7 & 55.3 & 69.9 & 71.8 & 73.2 & 67.6 \\
GPC \cite{jiang2021guided} & ICCV'2021 & 54.1 & 62.0 & 62.5 & 62.8 & 60.4 & - & - & - & - & - \\
IGNet \cite{ignet} & WACV'2024 & 49.0 & 61.3 & 63.1 & 64.8 & 59.6 & - & - & - & - & - \\
LiM3D \cite{lessismore} & CVPR'2023 & 58.4 & 62.2 & 63.1 & 63.6 & 61.8 & - & - & - & - & - \\
LASS3D \cite{li2025lass3d} & ECCV'2024 & 58.5 & 63.0 & 64.1 & 64.5 & 62.5 & - & - & - & - & - \\
DDSemi \cite{li2024density} & CVPR'2024 & \underline{59.3} & \underline{65.1} & \underline{66.3} & \underline{67.0} & \underline{64.4} & \underline{58.1} & \underline{70.2} & \underline{74.0} & \textbf{76.5} & \underline{69.7} \\ \hline
Ours (AIScene) & - & \textbf{61.2} & \textbf{66.3} & \textbf{67.4} & \textbf{67.9} & \textbf{65.7} & \textbf{60.2} & \textbf{72.3} & \textbf{75.0} & \underline{76.4} & \textbf{70.9} \\
\emph{Improvement} $\uparrow$ & - & \textcolor{lightblue}{+1.9} & \textcolor{lightblue}{+1.2} & \textcolor{lightblue}{+1.1} & \textcolor{lightblue}{+0.9} & \textcolor{lightblue}{+1.3} & \textcolor{lightblue}{+2.1} & \textcolor{lightblue}{+2.1} & \textcolor{lightblue}{+1.0} & \textcolor{lightblue}{-0.1} & \textcolor{lightblue}{+1.2} \\ \hline
\end{tabular}}
\vspace{-2mm}
\caption{Experimental results on the SemanticKITTI and nuScenes dataset compared with different SSL methods.
The results are reported with mIoU scores (\%). The \textbf{bold} and \underline{underline} scores respectively denote the best and second results.
}
\vspace{-4mm}
\label{tab:main_result}
\end{table*}

\subsection{Model Training}

Following the standard teacher-student mutual learning paradigm~\cite{tarvainen2017mean,lasermix,sohn2020fixmatch}, we construct a mini-batch of data with an equal composition of labeled and unlabeled scenes at each iteration. The unlabeled scenes are assigned with pseudo-labels by the teacher model and then train the student model jointly with labeled scenes. We use three types of loss functions to supervise the model training here, which is shown as follows:  
\begin{align} 
    &\ \mathcal{L}_{s}=\mathcal{L}_{seg} (\Phi_{\theta_\mathbbm{s}}(x_i^l), y_i^l), \\
    &\ \mathcal{L}_{l}=\mathcal{L}_{seg} (\Phi_{\theta_\mathbbm{s}}(x_i^{lpf}), y_i^{lpf}), \\
    &\ \mathcal{L}_{u}=\mathcal{L}_{seg} (\Phi_{\theta_\mathbbm{s}}(\hat{x}_i^{upf}), \hat{y}_i^{upf}), 
\end{align}
where $\mathcal{L}_{s}$, $\mathcal{L}_{l}$, and $\mathcal{L}_{u}$ denote the segmentation loss for the labeled scene $x_i^l$, augmented labeled scene $x_i^{lpf}$, and augmented unlabeled scene $\hat{x}_i^{upf}$, respectively, while $\mathcal{L}_{seg}$ is the original loss of the segmentation network $\Phi(\cdot)$. We sum up all the loss term to define the final training objective as:
\begin{align} \label{overall}
    \mathcal{L}=\mathcal{L}_{s}+\lambda_u\mathcal{L}_{u}+\lambda_l\mathcal{L}_{l}, 
\end{align}
where $\lambda_u$ and $\lambda_l$ are loss weights that balance the two loss terms. After that, we update the teacher model weights using the EMA strategy described in \cref{ema}.

\section{Experiments}
\subsection{Datasets and Evaluation Metrics}
\noindent\textbf{SemanticKITTI.}
Following state-of-the-art methods~\cite{lasermix,li2024density}, we evaluate our AIScene on the SemanticKITTI dataset~\cite{kitti}, which is a large-scale segmentation dataset consisting of LiDAR data derived from the popular KITTI dataset~\cite{old_kitti}.
It is composed of 22 sequences, with sequences 00-10 serving as the \emph{train} set (of which 08 as the \emph{val} set) and sequences 11-21 as the \emph{test} set. 
Moreover, all the point cloud scenes are collected from real-world rural and urban road environments in Germany, and the semantic class labels are mapped to 19 classes in our experiments.

\noindent\textbf{nuScenes.}
The nuScenes dataset~\cite{nuscene} is another well-known large autonomous driving dataset. 
This dataset is composed of 1000 sequences of 20s duration (850 for training and validation, 150 for testing), which are captured by the LiDAR sensor from Boston and Singapore.
Following the official splitting protocol, the \emph{train} and \emph{val} sets contain 28,130 and 6,019 point cloud scenes, respectively, including per-points annotation for 16 semantic classes.

\noindent\textbf{Evaluation Metrics.}
For both the SemanticKITTI and nuScenes datasets, we sample different ratios of annotated \emph{train} scenes to obtain semi-supervised data.
Meanwhile, we report the performance on the \emph{validation} set using the mean Intersection over Union (mIoU) metric~\cite{li2024density,lessismore}.

\subsection{Implementation Details}
We conduct experiments with 4$\times$ NVIDIA V100 GPUs with 32GB RAM, based on MMDetection3D~\cite{mmdet3d2020}. 
For the threshold $\tau_s$ for pseudo-label generation and point erasure, we experimentally find that 0.9 yields relatively satisfactory results.
We use the MinkowskiNet~\cite{choy20194d} and Cylinder3D~\cite{zhu2021cylindrical} as the backbone network due to its widely adopted in previous methods~\cite{lasermix,puy2024three,wang2024groupcontrast}.
To facilitate patch-based data augmentation, we restrict our focus to point clouds within the X and Y dimensions in the range of $[-50, 50]m$ for both the SemanticKITTI and nuScenes datasets. 
Similar to mainstream work, we also adopt the consistency loss proposed in MeanTeacher~\cite{tarvainen2017mean} to align the predictions of the teacher and student networks. Additionally, the weights of the consistency loss are assigned as $1\times10^3$ and $2\times10^3$ for the SemanticKITTI and nuScenes datasets, respectively.
The weights for $\lambda_u$ and $\lambda_l$ are set as 1 for both datasets. Our method excludes all forms of model ensemble and test-time augmentation during the evaluation.

\subsection{Main Results}
In this subsection, we compare our method with current SOTA semi-supervised point cloud segmentation methods (\eg, LaserMix \cite{lasermix}, LiM3D \cite{lessismore} and DDSemi \cite{li2024density}), using sub-sampling of the datasets under various partition protocols~\cite{li2024density}.
Specifically, we split the entire training set into labeled subsets with ratios of 1\%, 10\%, 20\%, and 50\% in the SemanticKITTI~\cite{kitti} and nuScenes dataset~\cite{nuscene}.
For further comparison, we also include several classical semi-supervised learning methods~\cite{tarvainen2017mean,zou2018unsupervised,chen2021semi} from the 2D image domain adapted to 3D segmentation. 

\begin{figure}[t]
    \centering
    \includegraphics[width=0.99\linewidth]{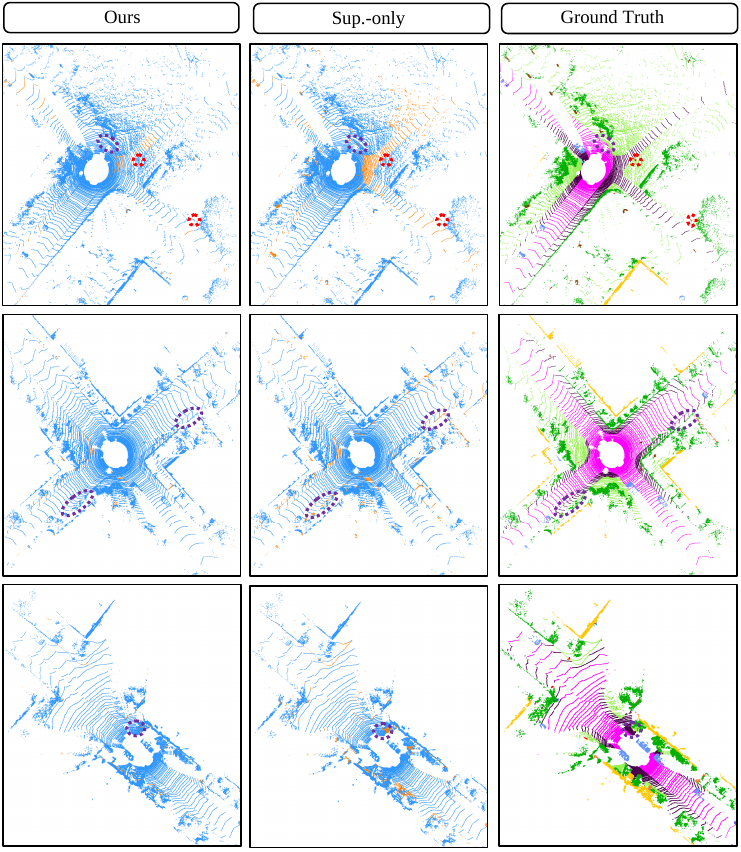}
    \caption{Qualitative comparison results from LiDAR bird's eye view on SemanticKITTI \emph{val} set. We set the point with \textcolor[rgb]{0.184,0.592,1}{correct} and \textcolor[rgb]{1,0.509,0.027}{incorrect} predicted semantic class in \textcolor[rgb]{0.184,0.592,1}{blue} and \textcolor[rgb]{1,0.509,0.027}{orange} for better visualization. Best viewed in color.}%
    \label{fig:vis_com}
    \vspace{-4mm}
\end{figure}

\cref{tab:main_result} summarizes the quantitative results on SemanticKITTI and nuScenes datasets. Our method consistently outperforms other comparative methods on both datasets, achieving average mIoU improvements of 1.3\% and 1.2\%, respectively. Notably, the proposed method demonstrates even greater gains on smaller sampling splits, such as 1\% and 10\%, which are particularly beneficial for the semi-supervised setting. Specifically, with a 1\% ratio, our method shows mIoU improvements of up to 1.9\% and 2.1\% on the two datasets, respectively, compared to the recently proposed SOTA method, DDSemi~\cite{li2024density}. These results also validate that our method is less sensitive to reduced labeled data sampling than other methods. Furthermore, we present qualitative results in \cref{fig:vis_com}. As indicated by the purple and red dashed circles, our method achieves better segmentation of boundary regions and small objects.

\subsection{Ablation Study}
In this subsection, We first validate the adaptability of our method to different segmentation models. We then examine the impact of intra-scene consistency through the point erasure strategy and explore inter-scene correlation by incorporating patch-based mixing and patch-based instance filling data augmentation.

\textbf{Adaptability with Different Segmentation Methods.}
Our AIScene is segmentor-agnostic and can be easily used to further improve off-the-shelf 3D fully-supervised segmentors under the semi-supervised setting. \cref{tab:abl_main1} shows that the proposed method can substantially boost the performance with Cylinder3D~\cite{zhu2021cylindrical} as the backbone, indicating the generalization ability of our framework. 
For consistency in comparison, the subsequent experiments are based on Cylinder3D unless otherwise specified.

\begin{figure*}[ht]
\setlength{\abovecaptionskip}{-0.0cm}
    \centering
    \includegraphics[width=\linewidth]{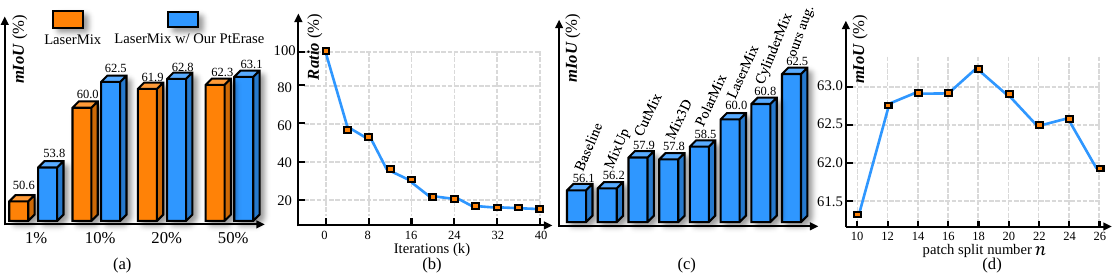}  % 16.5
    \put(-232,68){\fontsize{6pt}{6pt}\selectfont \rotatebox{65}{\cite{zhu2021cylindrical}}}
    \put(-220,65.5){\fontsize{6pt}{6pt}\selectfont \rotatebox{65}{\cite{zhang2017mixup}}}
    \put(-205,81.5){\fontsize{6pt}{6pt}\selectfont \rotatebox{65}{\cite{yun2019cutmix}}}
    \put(-192.5,78.5){\fontsize{6pt}{6pt}\selectfont \rotatebox{65}{\cite{nekrasov2021mix3d}}}
    \put(-176.5,90){\fontsize{6pt}{6pt}\selectfont \rotatebox{65}{\cite{xiao2022polarmix}}}
    \put(-162,102){\fontsize{6pt}{6pt}\selectfont \rotatebox{65}{\cite{lasermix}}}
    \put(-147.5,117){\fontsize{6pt}{6pt}\selectfont \rotatebox{65}{\cite{chen2024beyond}}}
    \caption{(a) Illustration of performance differences whether using point erasure strategy on the LaserMix~\cite{lasermix}. (b) Variation in the proportion of points removed by the point erasure strategy throughout the training process. (c) Comparison of different mixing-based techniques on SemanticKITTI dataset. (d) Results of mIoU against the number of patch split $n$ under the 10\% SemanticKITTI split.}
    \label{fig:consistency_multiscenes}
    \vspace{-4mm}
\end{figure*}

\begin{figure}[t]
    \centering
    \includegraphics[width=0.99\linewidth]{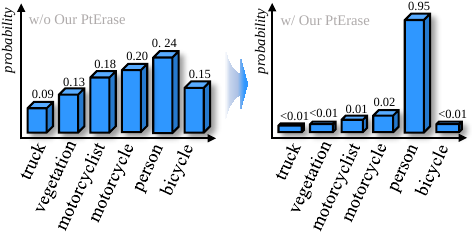}
    \vspace{-3mm}
    \caption{The left chart shows the initial output with distributed probabilities across multiple classes, indicating uncertainty in class prediction. Through our point erasure strategy, the right chart exhibits a distribution with a dominant class (``person'') having a significantly higher probability,}%
    \label{fig:probability}
    \vspace{-4mm}
\end{figure}

\textbf{Effectiveness of Point Erasure Strategy.} 
The 1\textsuperscript{st} row in \cref{tab:abl_main} reports the performance of the supervision-only Cylinder3D model~\cite{zhu2021cylindrical}, which serves as the baseline.
When PtErase is applied to the baseline, it effectively boosts performance by ensuring consistency during forward and backward propagation, thus addressing the limitations of the pseudo-label mechanism.
This consistency is achieved through a plug-and-play point erasure strategy, which can be easily integrated into methods relying on the pseudo-label mechanism. 
To further validate the importance of maintaining consistency during forward and backward propagation, we embedded the PtErase strategy into LaserMix~\cite{lasermix}. As illustrated in \cref{fig:consistency_multiscenes} (a), using the PtErase strategy improves performance across all ratios, underscoring the significance of intra-scene consistency and demonstrating the generalizability of our method. Moreover, \cref{fig:consistency_multiscenes} (b) shows that the proportion of points removed by the erasure strategy is initially high at the start of training, but gradually decreases as the network converges. Furthermore, We present a comparison of class probability distributions in \cref{fig:probability} to validate the effectiveness of the point erasure strategy. The results demonstrate that this strategy reduces classification ambiguity, enabling the model to better capture the features of the correct class.

\begin{table}[t]
\centering
\resizebox{\columnwidth}{!}{
\begin{tabular}{c|c|cccc|c}
\hline
 &  &  &  &  &  &  \\
\multirow{-2}{*}{Datasets} & \multirow{-2}{*}{Methods} & \multirow{-2}{*}{1\%} & \multirow{-2}{*}{10\%} & \multirow{-2}{*}{20\%} & \multirow{-2}{*}{50\%} & \multirow{-2}{*}{Avg.} \\ \hline \hline
 & Cylinder3D \cite{zhu2021cylindrical} & 45.4 & 56.1 & 57.8 & 58.7 & 54.5 \\
 & Ours (AIScene) & 54.5 & 63.3 & 63.7 & 64.3 & 61.5 \\
\multirow{-3}{*}{SemanticKITTI} & \emph{Improvement} & \textcolor{lightblue}{+9.1} & \textcolor{lightblue}{+7.2} & \textcolor{lightblue}{+5.9} & \textcolor{lightblue}{+5.6} & \textcolor{lightblue}{+7.0} \\ \hline
 & Cylinder3D \cite{zhu2021cylindrical} & 50.9 & 65.9 & 66.6 & 71.2 & 63.7 \\
 & Ours (AIScene) & 56.6 & 70.2 & 72.8 & 73.9 & 68.4 \\
\multirow{-3}{*}{nuScenes} & \emph{Improvement} & \textcolor{lightblue}{+5.7} & \textcolor{lightblue}{+4.3} & \textcolor{lightblue}{+6.2} & \textcolor{lightblue}{+2.7} & \textcolor{lightblue}{+4.7} \\ \hline
\end{tabular}}
\caption{Ablation study on the adaptability of our AIScene using Cylinder3D as the segmentation model.}
\vspace{-4mm}
\label{tab:abl_main1}
\end{table}

\begin{table}
\centering
\resizebox{\columnwidth}{!}{
\begin{tabular}{ccc|cccc|c}
\hline
\multicolumn{3}{c|}{Ours AIScene}                                                     & \multicolumn{4}{c|}{Splits on SemanticKITTI} & \multirow{2}{*}{Avg.} \\
 PtErase                   & MixPatch                    & InsFill                   & 1\%         & 10\%        & 20\%        & 50\%       &                       \\ \hline \hline
                         &                           &                           & 45.4        & 56.1        & 57.8        & 58.7       & 54.5                                      \\
 \checkmark &                           &                           & 51.7        & 59.8        & 61.8        & 60.4       & 58.4                                      \\
                           & \checkmark &                           & 51.3        & 61.4        & 60.7        & 60.5       & 58.5                                      \\
                           &                           & \checkmark & 52.3        & 61.7        & 60.1        & 61.1       & 58.8                                      \\
 \checkmark &                           & \checkmark & \textbf{54.6}        & 61.0        & 62.9        & 62.8       & 60.3                                      \\
 \checkmark & \checkmark &                           & 52.0        & 61.7        & 60.8        & 62.6       & 59.3                                      \\
                           & \checkmark & \checkmark & 50.5        & 62.5        & \textbf{63.8}        & 63.6       & 60.1                                      \\
 \checkmark & \checkmark & \checkmark & 54.5        & \textbf{63.3}        & 63.7        & \textbf{64.3}       & \textbf{61.5}                                      \\ \hline
\end{tabular}}
\caption{Ablation study of different components of our AIScene framework. PtErase refers to the point erasure strategy, while MixPatch and InsFill refer to the patch mix and instance filling data augmentation, respectively.} 
\vspace{-4mm}
\label{tab:abl_main}
\end{table}

\textbf{Effectiveness of Patch-based Data Augmentation.}
The patch-based data augmentation consists of MixPatch (patch-based mixing) and InsFill (instance filling).
In \cref{tab:abl_main}, the results in the 3\textsuperscript{rd} and 4\textsuperscript{th} rows indicate that using MixPatch or InsFill individually yields an average mIoU improvement of 4.0\% or 4.3\%, respectively, over the baseline.
This suggests that incorporating diverse semantic information across scenes benefits semi-supervised LiDAR segmentation. Furthermore, as shown in the 5\textsuperscript{th} and 6\textsuperscript{th} rows, combining MixPatch or InsFill with PtErase significantly enhances performance, highlighting the effectiveness of patch-based data augmentation. The combination of all three components yields substantial performance gains, as evidenced by the last row. Additionally, the patch pool and instance pool within the patch-based data augmentation facilitate the fusion of diverse semantic information across multiple scenes. 
To verify the effectiveness of interactions at the multi-scene level, we modify the pool implementation to allow interactions between only two scenes. 
The results in \cref{tab:abl_dual} indicate that multi-scene interactions yield superior performance compared to dual-scene interactions.

\textbf{Different Mixing Strategies.}
\cref{fig:consistency_multiscenes} (c) compares our patch-based data augmentation with existing mixing methods~\cite{nekrasov2021mix3d,xiao2022polarmix,yun2019cutmix}. MixUp~\cite{zhang2017mixup} shows a slight improvement over the baseline, likely due to its limited capacity for information transformation. Note that we reproduce CylinderMix in the LiDAR branch~\cite{chen2024beyond} for comparison. Other methods demonstrate considerable improvement over the baseline by enriching point cloud distributions. In contrast, our AIScene achieves the most substantial gains by effectively leveraging inter-scene correlation, highlighting its effectiveness in facilitating diverse semantic information generation for labeled and unlabeled scenes. This is achieved through an innovative design that incorporates a patch pool and an instance pool spanning multiple scenes.

\begin{table}[t]
\centering
\resizebox{\columnwidth}{!}{
\begin{tabular}{c|c|cccc|c}
\hline
 &  & \multicolumn{4}{c|}{Splits on Semantic KITTI} &  \\
\multirow{-2}{*}{Augmentation Type} & \multirow{-2}{*}{Methods} & 1\% & 10\% & 20\% & 50\% & \multirow{-2}{*}{Avg.} \\ \hline \hline
dual-scene level & Ours & 53.4 & 61.3 & 61.8 & 62.2 & 59.7 \\
multi-scene level & Ours & 54.5 & 63.3 & 63.7 & 64.3 & 61.5 \\
\emph{Improvement} & - & \textcolor{lightblue}{+1.1} & \textcolor{lightblue}{+2.0} & \textcolor{lightblue}{+1.9} & \textcolor{lightblue}{+2.1} & \textcolor{lightblue}{+1.8} \\ \hline
\end{tabular}}
\caption{
Comparison between dual-scene and multi-scene levels in the patch-based data augmentation of our AIScene.}
\label{tab:abl_dual}
\vspace{-4mm}
\end{table}

\textbf{Patch Split Number.}
As the patch split number $n$ plays a critical role in our framework, we further analyze the impact of varying $n$ in patch-based data augmentation, with the results presented in \cref{fig:consistency_multiscenes} (d). 
When $n$ is small, it can yield a certain level of enhancement, but it may fail to generate data sufficiently divergent from the original, thereby impeding the effective generation of diverse semantic information between labeled and unlabeled scenes. Conversely, when $n$ is too large, the generated scenes may deviate excessively from the original data, resulting in a loss of essential prior scene information. Empirically, we find that setting $n$ to 18 yields optimal results. Throughout this paper, we consistently use $n$ set to 18 unless explicitly stated otherwise.

\section{Conclusion}
This paper explores scene affinity for semi-supervised LiDAR semantic segmentation in driving scenes, named AIScene. Specifically, AIScene addresses intra-scene inconsistency through a point erasure strategy, which enhances the learning of pseudo-labeled points. Moreover, it employs a patch-based data augmentation method to mix multiple scenes at both scene and instance levels, guided by inter-scene correlation. This method enriches the semantic diversity of labeled scenes and complements pseudo-labeled scenes. Extensive experiments show that AIScene achieves state-of-the-art results on popular benchmarks. Additionally, the novel erasure strategy and augmentation method can function independently to facilitate semi-supervised LiDAR semantic segmentation.

\noindent\textbf{Acknowledgments.} 
This work is supported by the National Natural Science Foundation of China (No.62325111, U22B2011, 62271354).

{
    \small
    \bibliographystyle{ieeenat_fullname}
    \bibliography{main}
}

\end{document}